\documentclass{article}
\usepackage{ijcai21}

\usepackage{times}
\usepackage{soul}
\usepackage{url}
\usepackage[hidelinks]{hyperref}
\usepackage[utf8]{inputenc}
\usepackage[small]{caption}
\usepackage{graphicx}
\usepackage{booktabs}
\usepackage{algorithm}
\usepackage{algorithmic}
\usepackage{amsmath,amsfonts,bm}
\usepackage{calc,amsfonts,amssymb,amsmath,bm,theorem,epstopdf,nicefrac}
\usepackage{psfrag,float}

\DeclareMathOperator*{\minimize}{\rm minimize}

\newcommand{\X}{\boldsymbol{X}}

\newcommand{\x}{\boldsymbol{x}}

\def\eqref#1{eq.~(\ref{#1})}

\def\1{\bm{1}}

\def\vx{{\bm{x}}}
\def\vy{{\bm{y}}}
\def\vz{{\bm{z}}}

\def\mA{{\bm{A}}}

\def\mM{{\bm{M}}}

\def\mS{{\bm{S}}}

\def\mU{{\bm{U}}}
\def\mV{{\bm{V}}}
\def\mW{{\bm{W}}}
\def\mX{{\bm{X}}}
\def\mY{{\bm{Y}}}
\def\mZ{{\bm{Z}}}

\DeclareMathAlphabet{\mathsfit}{\encodingdefault}{\sfdefault}{m}{sl}
\SetMathAlphabet{\mathsfit}{bold}{\encodingdefault}{\sfdefault}{bx}{n}
\newcommand{\tens}[1]{\bm{\mathsfit{#1}}}

\def\tZ{{\tens{Z}}}

\def\gP{{\mathcal{P}}}
\def\gQ{{\mathcal{Q}}}

\newtheorem{Prop}{Proposition}

\theorembodyfont{\rmfamily}

\newcommand{\R}{\mathbb{R}}

\DeclareMathOperator*{\argmin}{arg\,min}

\newcommand{\etal}{\textit{et al.}}

\usepackage{color}
\definecolor{orange}{RGB}{255,107,0}

\graphicspath{ {./imgs/} }

\usepackage{subfigure}
\usepackage{multirow}

\title{Stochastic Block-ADMM for Training Deep Networks}

\author{
Saeed Khorram$^1$
\and
Xiao Fu$^1$\and
Mohamad H. Danesh$^{1}$\and
Zhongang Qi$^2$\and
Li Fuxin$^1$
\affiliations
$^1$Oregon State University\\
$^2$Applied Research Center, PCG, Tencent \\
\emails
$^1$\{khorrams, xiao.fu, daneshm, lif\}@oregonstate.edu,
$^2$zhongangqi@tencent.com
}

\begin{document}

\maketitle

\begin{abstract}

In this paper, we propose Stochastic Block-ADMM as an approach to train deep neural networks in batch and online settings. Our method works by splitting neural networks into an arbitrary number of blocks and utilizes auxiliary variables to connect these blocks while optimizing with stochastic gradient descent. This allows training deep networks with non-differentiable constraints where conventional backpropagation is not applicable. An application of this is supervised feature disentangling, where our proposed DeepFacto inserts a  non-negative matrix factorization (NMF) layer into the network. Since backpropagation only needs to be performed within each block, our approach alleviates vanishing gradients and provides potentials for parallelization. We prove the convergence of our proposed method and justify its capabilities through experiments in supervised and weakly-supervised settings.

\end{abstract}

\section{Introduction} \label{sec:intro}

Deep Neural Networks (DNNs) are highly non-convex functions with ill-conditioned Hessians and are believed to have multiple local minima and saddle points. Most networks are trained with 
Stochastic Gradient Descent (\textit{SGD}) and its adaptive learning rate variants \textit{e.g.,} \textit{Adam} \cite{kingma2014adam} are used to optimize the DNNs with backpropagation. Although these approaches have been the most successful, they suffer from issues such as vanishing gradients in deep layers, a significant memory footprint for storing the gradients, and difficulty to parallelize across layers because backpropagation has to be done sequentially\cite{taylor2016training}. In addition, in the presence of non-differentiable layers, conventional backpropagation training cannot be applied.

Alternating Direction Method of Multipliers (ADMM) is a simple yet powerful approach that decouples optimization variables and optimizes the augmented Lagrangian in a primal-dual scheme. 
It has shown promise in solving certain families of non-convex problems \cite{wang2019global,huang2018mini}.
Recently, optimization of the neural networks with such alternating direction techniques has gained rising attention \cite{zeng2018global,zeng2019convergence,zhang2017convergent,gu2018fenchel,askari2018lifted} which would potentially avoid the disadvantages of the SGD and introduce beneficial properties such as \textit{fast(er)} convergence, ease of \textit{parallelization} and \textit{distributed} training, and being able to enforce additional (non-differentiable) constraints on the DNN tensors.

Despite their advantages, there are several reasons ADMM-like methods are not widely used in DNN training. The performance of these methods is usually not as good as conventional backpropagation with SGD variants, the algorithms are usually batch mode which directly restricts the number of trainable parameters and training data as well, updates are in closed-from which prohibits the use of complicated architectures while being memory intensive, \textit{etc.} Further, existing ADMM-like methods have restrictive assumptions in the architecture of the network which prohibits the extension to non-trivial networks such as ResNets \cite{he2016deep}. Work of \cite{taylor2016training} is of this kind which, despite the parallelization capabilities introduced by ADMM, the size of the training data is linearly limited by the number of cores. 

In this paper, we propose Stochastic Block-ADMM which addresses the aforementioned issues. Stochastic Block-ADMM separates DNN parameters into an arbitrary number of blocks and uses stochastic gradients to update each block. The error signals are passed between the blocks by introducing auxiliary variables at the splitting points. We present both batch and online versions of the Stochastic Block-ADMM which can be extended to settings where computational resources are limited, data is constantly changing such as in reinforcement learning or training with data augmentation techniques. We provide a convergence proof for the proposed approach and verify its performance on several deep learning benchmarks.

An ADMM formulation of deep networks also allows us to add additional \textit{non-differentiable} constraints to the learning problem. In this paper, 
we explore the problem of supervised feature disentanglement by inserting non-negative factorization layers into the network. Nonnegative Matrix Factorization (NMF) has been shown to generate sparse and interpretable representations due to the non-negative constraints over the factorization matrices \cite{lee1999learning}. Jointly training an NMF decomposition with deep learning adds non-differentiable non-linearity and \textit{cannot} be addressed by the conventional backpropagation with SGD algorithms. We show results training these networks via ADMM and their performance on a supervised feature disentanglement benchmark.

In summary, our paper makes the following contributions:
\begin{itemize}
    \item We propose Stochastic Block-ADMM for training deep networks. This improves over previous ADMM approaches (in training deep networks) which only work in batch setting.
    \item We propose an online variant of the Stochastic Block-ADMM for further efficiency in computations. 
    \item We prove the convergence of the proposed Stochastic Block-ADMM algorithm.
    \item We propose DeepFacto, which jointly trains a non-negative matrix factorization layer with a deep network using ADMM, and show its capability in supervised feature disentanglement.
\end{itemize}


\section{Related Work}\label{sec:rel_work}

Alternating Direction Method of Multipliers (ADMM) has shown promise in solving optimization problems, especially in large-scale and data-distributed machine learning applications. The power of ADMM comes from its decomposition of the augmented Lagrangian into simpler loosely-coupled sub-problems which enables it to solve each sub-problem in an efficient and potentially parallel manner.
ADMM extensions for non-convex problems have been recently proposed which are more suitable for large data sets and more complicated problems \cite{wang2019global,huang2018mini}.


A recent line of research has focused on training DNNs using optimization techniques that decompose the training into smaller subproblems, including Block Coordinate Descent (BCD) and ADMM. On the BCD algorithms, \cite{carreira2014distributed} was the earliest to propose training a DNN in a distributed setting by formulating it as a constrained optimization problem. Further, \cite{zeng2018global,zhang2017convergent,askari2018lifted,gu2018fenchel} lifted the non-convex activations (e.g. ReLU) and formulating the DNN training as a multi-convex problem and solved it using BCD and \cite{choromanska2018beyond} proposed an online method for training DNNs.

On the other hand, \cite{taylor2016training} proposed a batch gradient-free algorithm for training neural networks using a variant of ADMM. However, due to the closed-form update of all the parameters, the proposed method has limitations (\textit{e.g.} only capable of using simple losses such as Hinge loss and MSE), and cannot be further extended into more complex problems and larger datasets. However, the scope of
\cite{zhang2016efficient} is limited to a specific application and no convergence proof is presented.

\cite{gotmare2018decoupling} splits DNNs into blocks and trained them separately by introducing gluing variables. This is very close to ADMM, but it did not use the dual variables common in ADMM and did not present a convergence proof for their method.
Recently, \cite{wang2019admm,zeng2019convergence} have provided convergence analysis of ADMM (to a stationary point) in deep learning by linearly approximating the non-linear constraints in the DNN training problem. However, their work did not address stochastic gradients as in our work.

Non-negative Matrix Factorization (NMF) imposes  non-negativity constraints over the factors, hence can lead to more interpretable decompositions than methods such as Principle Component Analysis (PCA)~\cite{lee1999learning,liu2011constrained}.  
\cite{collins2018deep} applied NMF over convolutional activations which has shown interpretable and coherent behavior over image parts. However, in their work, NMF was applied post-hoc over pre-trained CNN activations. There is no guarantee that the disentanglement is faithful to the underlying mechanism of the DNN. 
To the best of our knowledge, NMF layers jointly trained with a deep neural network have not been studied in the past.


\section{Method} \label{sec:method}

There were many hurdles in using ADMMs for deep learning --- the global convergence proof of the ADMM \cite{deng2016global} assumes that the optimization objective is deterministic and the global solution is calculated at each iteration of the cyclic parameter updates.
This typically requires matrix inversion and makes standard ADMM computationally expensive thus impractical for training of many large-scale optimization problems. To see a formulation of standard ADMM for training DNNs refer to the supplementary materials \ref{sec:admm_nn}.  

In this section, we present stochastic Block-ADMM which does not require global solution as well as an online version which further reduces the communication load. We prove the convergence of these algorithms in Sec. \ref{sec:convergence} and present its application in supervised disentanglement in Sec.~\ref{sec:deepfacto}.

\begin{figure*}[t]
\begin{center}
\subfigure[] { \label{fig:block_admm}
\includegraphics[width=0.75\linewidth]{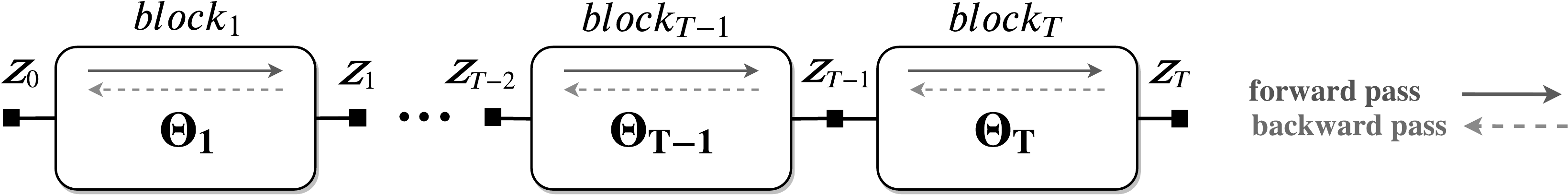}
}
\subfigure[] { \label{fig:block}
\includegraphics[width=0.15\linewidth]{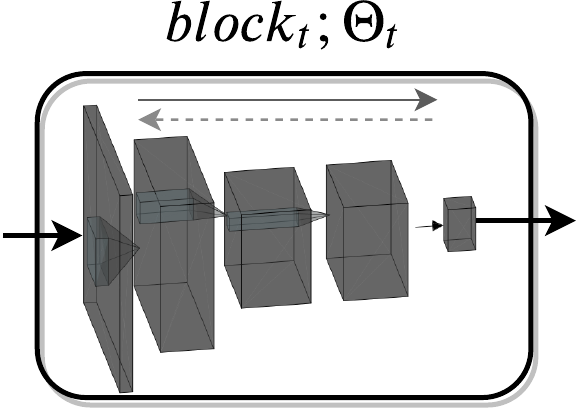}
}
\end{center}
\caption{\small a) General Architecture for training DNNs proposed in Stochastic block-ADMM. b) A few differential layers selected from a parent network are stacked inside a block. The parameters $\Theta_t$ are updated by SGD in a forward-backward pass.}
\end{figure*}


\subsection{Stochastic Block-ADMM}\label{sec:block_admm}

In this section, we introduce a novel variant of ADMM for training DNNs, the stochastic block-ADMM. We first split the conventional multi-layer network architectures into an arbitrary number of \emph{blocks}, each containing only a part of the network. To make the parameters of each block independent from its neighbors, \emph{decoupling variables} \{$\mZ_t, \; t=1, \dots, T$\} are introduced as shown in Fig.~\ref{fig:block_admm}. These variables pass the information forward and backward in the architecture to train blocks in a cyclic manner until consensus is reached. Each $block_t$  consists of one or multiple differentiable layers (e.g., convolutional layers, activation layers, etc.) that are detached from the rest of the network via coupling variables. Denote the set of all learnable parameters of each $block_t$ as $\Theta_t$. As an example, a $block_t$ wrapping multiple layers can be seen in Figure \ref{fig:block}. Our formulation is:
\begin{align} \label{eq:ourformulation}
	\minimize_{ {\bm \Theta}, \mathcal{Z}}\; &\mathcal{J}\left(\mY, \mZ_{T} \right) 
	 \\
 {\rm subject~to} ~ &\bm Z_t = \mathrm{block}_{\bm \Theta_t}(\bm Z_{t-1}), \quad \mZ_{0} = \mX \nonumber
\end{align}
where ${\bf \Theta} = \{\Theta_t\}_{t=1}^{T} \text{and } \mathcal{Z} = \{\mZ_t\}_{t=1}^{T}$. $\mathcal{J}$ is the desired cost to be minimized (\textit{e.g.}, cross-entropy loss), $T$ is the total number of blocks, $\mX = \{ \vx_1,\dots, \vx_N \} \in  \R^{M \times N}$ is the input data, and $\mY = \{\vy_1,\dots, \vy_N \} \in \R^{C \times N}$ is the target label -- for $C$ classes. Note that the number of blocks $T$ can be different than the number of layers in the network $L$.

To train DNNs with this new approach, 
we would have the following augmented Lagrangian minimization problem to enforce the equality constraints needed for training,
\begin{align} \label{eq:block_admm_unconstrained}
	\min_{ {\bf \Theta}, \mathcal{Z}} \; &\mathcal{J}\left(\mY, \mZ_{T} \right) 
	+ \sum_{t=1}^{T} \frac{\beta_t}{2} \| \mZ_t - \mathrm{block}_{\Theta_t}(\mZ_{t-1}) + \mU_t\|_F^2 \nonumber \\
	& {\rm subject~to} \quad \mZ_{0} = \mX 
\end{align}
where $\beta_t$ and $\mU_t$ are the (scaled) step size  and the Lagrange multiplier corresponding to the $t$-th Block, respectively. Our proposed Stochastic block-ADMM method for training problem (\ref{eq:block_admm_unconstrained}) is presented in Algorithm \ref{alg:blockadmm}. 
$\zeta_t$ and $\eta_t$ are the learning rates in each update step for $\mZ_t$ and $\Theta_t$, respectively. Similar to training conventional neural networks, each block is updated by first going in a forward pass through the block and update the parameters using back-propagation. Update of the block parameters $\Theta_{t}$ is done using mini-batch stochastic gradient descent or Adam. The same goes for the decoupling variables $\mZ_t$. Note, in each cycle of the parameter update in Algorithm \ref{alg:blockadmm}, all the samples of $\mZ$ are updated, while $\Theta_{t}$ is updated stochastically. In addition, due to non-convexity of primal sub-problem (Eq. \ref{eq:primal}), one can perform the primal updates for multiple steps. 
In Algorithm \ref{alg:blockadmm}, we take the reverse order for updating  the decoupling variables $\mZ_t$, which we have empirically found more efficient, as analogous to backpropagation where gradient flows backwards as well.

Note that in this formulation, 
backpropagation stops at each auxiliary variable $\mZ_t$ . Hence, our method can readily mitigate the long-known vanishing gradient problem by splitting a conventional DNN into arbitrary sized blocks. 
During testing time, one could follow Eq. (\ref{eq:block_admm_unconstrained}) to solve an optimization problem. But in practice, it suffices to use a straight-through estimator by removing the decoupling variables and simply pass the output of each layer to the next, equivalent of doing a forward pass in a conventional DNN. 

\begin{algorithm}[htb]
   \caption{Stochastic Block-ADMM}
   \label{alg:blockadmm}
\begin{algorithmic}
   {\STATE \scalebox{1}{\bfseries Input:} data $\mX$, labels $\mY$}
   \STATE  \scalebox{1}{{\bfseries Params:} $\beta_t >0, \; \zeta_t >0, \eta_t >0$ }
   \STATE  {\bfseries Define:} \scalebox{0.80}{ $\mathcal{T}({\mZ_{t}, \mZ_{t-1}, \mU_{t}, \Theta_t}) = \frac{\beta_t}{2} \| \mZ_t - \mathrm{block}_{\Theta_t}(\mZ_{t-1}) + \mU_t\|_F^2$ }
   \STATE  \scalebox{1}{{\bfseries Initialize:} $\{{\Theta_t^0}\}_{t=1}^{T}, \{ \mU_t^0\}_{t=1}^{T} ,\; k \leftarrow 0$ }
   \STATE  \scalebox{1}{{\bfseries Initialize:} $\{\mZ_t\}_{t=1}^{T}$ in a forward pass. }
   \REPEAT
   \STATE \scalebox{1}{ $\mZ_{T}^{k+1} \leftarrow \mZ_{T}^{k} - \zeta_T \nabla_{\mZ_{T}^k} ( \mathcal{J}\left(\mY_{i}, \mZ_{T}^{k} \right)$ }
   \STATE \scalebox{1}{$ + \mathcal{T}({\mZ_{T}^k, \mZ_{T-1}^k, \mU_{T}^k, \Theta_L^k})) \;  $}
   \FOR{$t=T-1$ {\bfseries to} $1$}
   \STATE \scalebox{1}{ $\mZ_{t}^{k+1} \leftarrow\mZ_{t}^{k} - \zeta_t \nabla_{\mZ_{t}^k} ( \mathcal{T}({\mZ_{t}^k, \mZ_{t-1}^k, \mU_{t}^k, \Theta_{t}^k})$} \\
   \STATE  \scalebox{1}{$+ \mathcal{T}({\mZ_{t+1}^{k+1}, \mZ_{t}^k, \mU_{t+1}^k, \Theta_{t+1}^k})) \; $}
   \ENDFOR
   \FOR{$t=1$ {\bfseries to} $T$}
   \STATE \scalebox{0.9}{${\Theta_t}^{k+1} \leftarrow {\Theta_t}^{k} - \eta_t  \nabla_{\Theta_t} \mathcal{T}({\mZ_{t,i}^{k+1}, \mZ_{t-1,i}^{k+1}, \mU_{t,i}^k, \Theta_{t}^k}),$}
   \STATE \scalebox{0.9}{$draw \; i \subset \{1,\dots,N\} \; $}
   \STATE \scalebox{1}{$\mU_t^{k+1} \leftarrow \mU_t^{k} + \mZ_{t}^{k+1} - \mathrm{block}_{\Theta_t}^{k+1}(\mZ_{t-1}^{k+1})$}
   \ENDFOR
   \UNTIL{some stopping criterion is reached.}
\end{algorithmic}
\end{algorithm}

\subsection{Online Stochastic Block-ADMM}\label{sec:onlineadmm}

The stochastic block-ADMM formulation in section \ref{sec:block_admm} is still a batch mode algorithm, in the sense that the entire training set is updated at once. This imposes restrictions on the size of the input and the number of parameters in the network when limited resources are available. 
Also, it does not readily accommodate to settings where data is constantly changing, such as data augmentation on the input or reinforcement learning. To overcome such limitations, we propose an \textit{online} variant of the stochastic block-ADMM in Algorithm \ref{alg:online_admm} which alternatively solves the unconstrained problem, 
\begin{align}\label{eq:scalar_dual}
	\min_{ {\bf \Theta}, \mathcal{Z}} \; &\mathcal{J}\left(\vy, \vz_{T} \right) 
	+ \sum_{t=1}^{T} \frac{\beta_t}{2} \big( \|\vz_t - \mathrm{block}_{\Theta_t}(\vz_{t-1}) \|_F^2 + u_t\big) \nonumber\\
	& {\rm subject~to} \quad \vz_{0} = \vx 
\end{align}
Although similar to the Eq. (\ref{eq:block_admm_unconstrained}), the dual variable in the online Block-ADMM is a \textit{scalar}. The benefits of this are two-folded: First, this substantially reduces the memory size needed for storing the dual variables as the optimization proceeds. Second, this considerably reduces the variance in the gradient induced by re-initializing the auxiliary variables $\vz_{\ell,i}$ when updating the block parameters at each iteration. 


\begin{algorithm}[htb]
   \caption{Online Stochastic Block-ADMM }
   \label{alg:online_admm}
\begin{algorithmic}
   {\STATE \scalebox{1}{\bfseries Input:} data $\mX$, labels $\mY$}
   \STATE  \scalebox{1}{{\bfseries Params:} $\beta_t >0, \; \zeta_t >0, \eta_t >0$ }
   \STATE  {\bfseries Define:} \scalebox{0.8}{ $\mathcal{T}({\vz_{t}, \vz_{t-1}, u_{t}, \Theta_t}) = \frac{\beta_t}{2} (\| \vz_t - \mathrm{block}_{\Theta_t}(\vz_{t-1}) \|_2 + u_t)^2$ }
   \STATE  \scalebox{1}{{\bfseries Initialize:} $\{{\Theta_t^0}\}_{t=1}^{T}, \{ u_t^0\}_{t=1}^{T} ,\; k \leftarrow 0$ }
   \REPEAT 
   \FOR{$(\vx_i, \vy_i) \text{\bfseries in} (\mX,\mY)$}
   \STATE \scalebox{1}{{\bfseries Initialize:} $\{\vz_{t,i}\}_{t=1}^{T}$ in a forward pass $(\vz_{0,i} = \vx_i)$.}
   \STATE \scalebox{1}{$\vz_{T,i} \leftarrow \vz_{T,i} - \zeta_T \nabla_{\vz_{T,i}} ( \mathcal{J}\left(\vy_{i}, \vz_{T,i} \right)$ }
   \STATE \scalebox{1}{$+\mathcal{T}({\vz_{T}, \vz_{T-1}, u_{T}^k, \Theta_T^k})) \;  $}
   \FOR{$t=T-1$ {\bfseries to} $1$}
   \STATE \scalebox{1}{ $\vz_{t,i} \leftarrow\vz_{t,i} - \zeta_t \nabla_{\vz_{t,i}} ( \mathcal{T}({\vz_{t,i}, \vz_{t-1,i}, u_{t}^k, \Theta_{t}^k})$} \\
   \STATE  \scalebox{1}{$+ \mathcal{T}({\vz_{t+1,i}, \vz_{t,i}, u_{t+1}^k, \Theta_{t+1}^k})) \; $}
   \ENDFOR
   \FOR{$t=1$ {\bfseries to} $T$}
   \STATE \scalebox{1}{${\Theta_t}^{k+1} \leftarrow {\Theta_t}^{k} - \eta_t  \nabla_{\Theta_t} \mathcal{T}({\vz_{t,i}, \vz_{t-1,i}, u_{t}^k, \Theta_{t}})$}
   \STATE \scalebox{1}{$u_t^{k+1} \leftarrow u_t^{k} + \| \vz_{t}^{k} - \mathrm{block}_{\Theta_t^{k+1}}(\vz_{t-1,i})\|_2$}
   \ENDFOR
   \ENDFOR
   \UNTIL{some stopping criterion is reached.}
\end{algorithmic}
\end{algorithm}

\subsection{Convergence of the Algorithm}\label{sec:convergence}
Let us consider the following general problem:
\begin{align}\label{eq:main}
	\minimize_{\mathcal{Z},\bf \Theta} & f(\mathcal{Z})\\
	{\rm subject to} & h(\mathcal{Z},\bm \Theta)=\bm 0,\nonumber
\end{align}
where $\mathcal{Z}$ and $\bf \Theta$ are as defined in Sec.~\ref{sec:block_admm}, and $f(\cdot)$ represents the training objective, and $h(\cdot)$ represents the layer coupling equalities as in \eqref{eq:ourformulation}.
We also assume that both $f(\cdot)$ and $h(\cdot)$ are differentiable functions. Note that both $f$ and $h$ can be non-convex.

Let us consider the following augmented Lagrangian:
\[          {\cal L}_{\rho_k}(\mathcal{Z},{\bf \Theta},{\bm \lambda})=f(\mathcal{Z}) + \langle \bm \lambda, h(\mathcal{Z},{\bf \Theta})\rangle + \frac{1}{2\rho_k}\|h(\bm Z,\bf \Theta)\|_2^2,  \]
where $\bm \lambda$ collects all the dual variables $\bm U_1,\ldots,\bm U_T$ that correspond to different layers. The standard primal-dual updates can be summarized as follows:
\begin{subequations}\label{eq:stopdd}
\begin{align}
    (\bm Z^{k+1},\Theta^{k+1}) &\leftarrow  \arg\min_{\mathcal{Z},\bf \Theta}  {\cal L}_{\rho_k}(\mathcal{Z},\bf \Theta,\bm \lambda^k), \label{eq:primal}\\
    \bm \lambda^{k+1} &\leftarrow \bm \lambda^k + \frac{1}{\rho_k}h(\bm Z^{k+1}, \Theta^{k+1}),
\end{align}
\end{subequations}
We employ the trick in \cite{shi2017penalty} for adaptively adjusting the parameter $\rho_k$. We assume that $\rho_k$ is adjusted by
\begin{align}\label{eq:rho}
    \rho_{k+1} \leftarrow \begin{cases}  \rho_k,&\quad \|h(\bm Z^{k},\bm \Theta^k)\|\leq \eta_k,\\
                                         c\rho_k,~0<c<1,&\quad {\rm o.w.}
    \end{cases}
\end{align}
where $\eta_k$ for $k=1,2,\ldots$ is a pre-specified sequence that bounds the equality-enforcing error.

Our analysis shows the following convergence result:
\begin{Prop}\label{prop:convergence}
    Assume $h({\cal Z},\bm \Theta)=\bm 0$ satisfies the Robinson's condition.
    Also assume for each update in \eqref{eq:primal}, the sub-problem solution solved by stochastic alternating optimization satisfies
   \begin{equation}
       \mathbb{E}\left[ \left\| {\cal G}(\x^k) \right\|^2\right]\leq \varepsilon_k, ~ \mathbb{V}\left[  {\cal G}(\x^k) \right]\leq \sigma_k^2,
   \end{equation}  
   where $\x=({\cal Z},\bm \Theta)$ is a vector that collects all the optimization variables and ${\cal G}(\x^k)$ collects the stochastic gradients that we used for updating $({\cal Z},\bm \Theta)$.
   Assume that the stochastic gradient for the primal update is unbiased, i.e.,
   \begin{equation}\label{eq:unbiasedness}
       \mathbb{E}[{\cal G}(\x^k)] = \nabla {\cal L}_{\rho_k}(\x_k),~\forall k. 
   \end{equation}         
   Then, every limit point of the solution sequence produced by the algorithm in \eqref{eq:stopdd} converges to a KKT point of the problem in~\eqref{eq:main}, if $\eta_k\rightarrow 0$, $\sigma_k^2 \rightarrow 0$ and $\varepsilon_k\rightarrow 0$.	
\end{Prop}
The proof for Proposition~\ref{prop:convergence} is presented in the supplementary materials \ref{sec:proof}.
Proposition~\ref{prop:convergence} asserts that the algorithm converges to a KKT point under some conditions. 
   There are a number of remarks regarding implementation.
   To begin with, the condition $\varepsilon_k\rightarrow 0$ means that the primal problem needs to be solved more and more accurately when $k$ grows, in terms of approaching the stationary point of the sub-problem using block stochastic gradient. This can be achieved via gradually increasing the number of iterations for the primal updates. Note that stochastic block gradient can provably attain $\mathbb{E}[\|{\cal G}(\X^k)\|^2]\leq \varepsilon_k$; see \cite{xu2015block}. 
\subsection{DeepFacto: Factorization of DNN Activations }\label{sec:deepfacto}

Here, we investigate a task for supervised disentanglement, which can provide insights for explaining DNNs to humans. Supervised disentanglement aims to find disentangled factors that decide the CNN output, yet are human-understandable and distinct from each other. One approach to learn a disentangled representation is through adding  non-negative matrix factorization (NMF)\cite{lee1999learning} layers to the network \cite{collins2018deep}. Note that NMF imposes non-differentiable constraints into the network where conventional end-to-end training using backpropagation would not be applicable. Hence, prior work were mostly running NMF after the training, where the network might have already learned highly entangled features. In this work, aided with our stochastic block-ADMM, we attempt to perform training with NMF layers in the intermediate layers of DNNs.

\begin{figure}[t!]
\begin{center}
\centerline{
\includegraphics[width=1 \columnwidth]{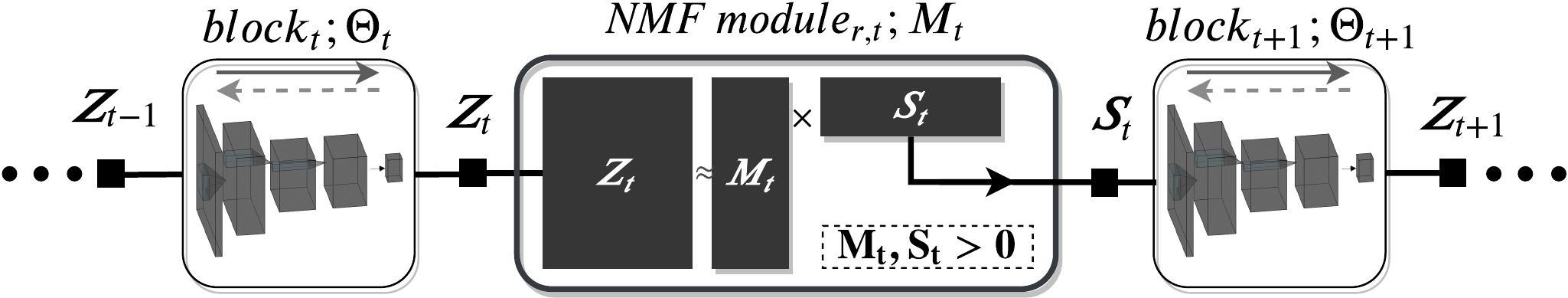}
}
 \caption{General architecture for Deepfacto: an NMF module with rank $r$ is added in the middle of two arbitrary blocks. Note, only $\mS_t$ is passed to the next blocks.}
\label{fig:deepfacto}
\end{center}
\end{figure}

Figure \ref{fig:deepfacto} shows an \emph{NMF module} with \emph{rank $r$} incorporated between two arbitrary neighboring blocks. The output from the $block_t$ is factorized into $\mM_t$ and $\mS_t$, namely, the basis and score matrices. In this configuration, only the score matrix $\mS_t$ is passed to the next blocks. The score matrix is low-rank, sparse and non-negative hence can possibly represent features that are more disentangled than the original network. 
Exploring this architecture is one attempt of us in making deep networks more explainable to humans. Humans would not be able to interpret conventional deep network weights which are both positive and negative and sometimes cancels out each other. The sparse and non-negative feature from NMF would be much more preferable to interpret~\cite{collins2018deep}.

However, the NMF module breaks the gradient path from $\mS_t$ to $Z_t$, hence conventional backpropagation would not be applicable in this problem. We extend the ADMM framework (\ref{eq:block_admm_unconstrained}) into having non-negative factorization constraints over its activations and formulate the following optimization problem:
\begin{eqnarray} \label{eq:block_admm_nmf}
	\min_{ {\bf \Theta}, \mathcal{Z}, \mS, \mM} \; &\mathcal{J}\left(\mY, \mZ_{T} \right) \nonumber \\
	+ & \sum_{k=1, k\neq t+1}^{T} \frac{\beta_k}{2} \| \mZ_k - block_k(\mZ_{k-1}) + \mU_k\|_F^2 \nonumber \\
	+ & \frac{\beta_{t+1}}{2} \| \mZ_{t+1} - block_{t+1}(\mS_{t}) + \mU_{t+1}\|_F^2 \nonumber \\
	+ &  {\frac{\gamma_t}{2} \| \mZ_t - \mM_t \mS_t + \mV_t\|_F^2} \nonumber \\
	  & {\forall i,j} \;  \mM_{\ell,ij} \ge 0,\; \mS_{\ell,ij} \ge 0 
\end{eqnarray}
where $\gamma_t$ is the step-size and $\mV_t$ is the corresponding multipliers to enforce the matrix factorization equality $\mZ_t = \mM_t \mS_t$. The NMF module adds a nonconvex term to the optimization. However, in the alternating optimization scheme, while keeping either $\mM_t$ or $\mS_t$ constant, solving for the other term would reduce to a normal convex least-squares problem. The rest of the updates are the same as in section \ref{sec:block_admm}. Note that, trivially to not change the input dimension of the next block after the NMF module, one can simply add an affine layer to increase the dimensions without changing the formulation.

At testing time, one only needs to perform a non-negative projection since the basis matrix $M$ will be given, which can be solved using a convex solver such as LBFGS. Note that for simplicity, we only formulated adding \emph{one} NMF module in the middle of the blocks. This can be simply extended to as many NMF modules as needed in the architecture.

\section{Experiments} \label{sec:exp}

All the experiments are run on a machine with a single NVIDIA GeForce RTX 2080 Ti GPU. The results presented for each of the following experiments are selected from their best performance after grid search over the hyper-parameters, both for our method and the baselines. 
Each algorithm is ran five times with different initialization and the average test set accuracy is reported. The shaded area corresponds to $\pm1$ standard deviation. We will make our code available online. 

\subsection{Supervised Deep Network Training}\label{exp:conv}
In this section, we present the experiment results from training conventional neural networks in a supervised setting on the MNIST, Fashion-MNIST, and CIFAR-10 datasets. For experiments results on Fashion-MNIST and CIFAR-10, see supplementary materials \ref{sec:sup_train}. 

\subsubsection{MNIST}\label{exp:mnist}
\begin{figure}[ht]
\begin{center}
\centerline{
\includegraphics[width=\columnwidth]{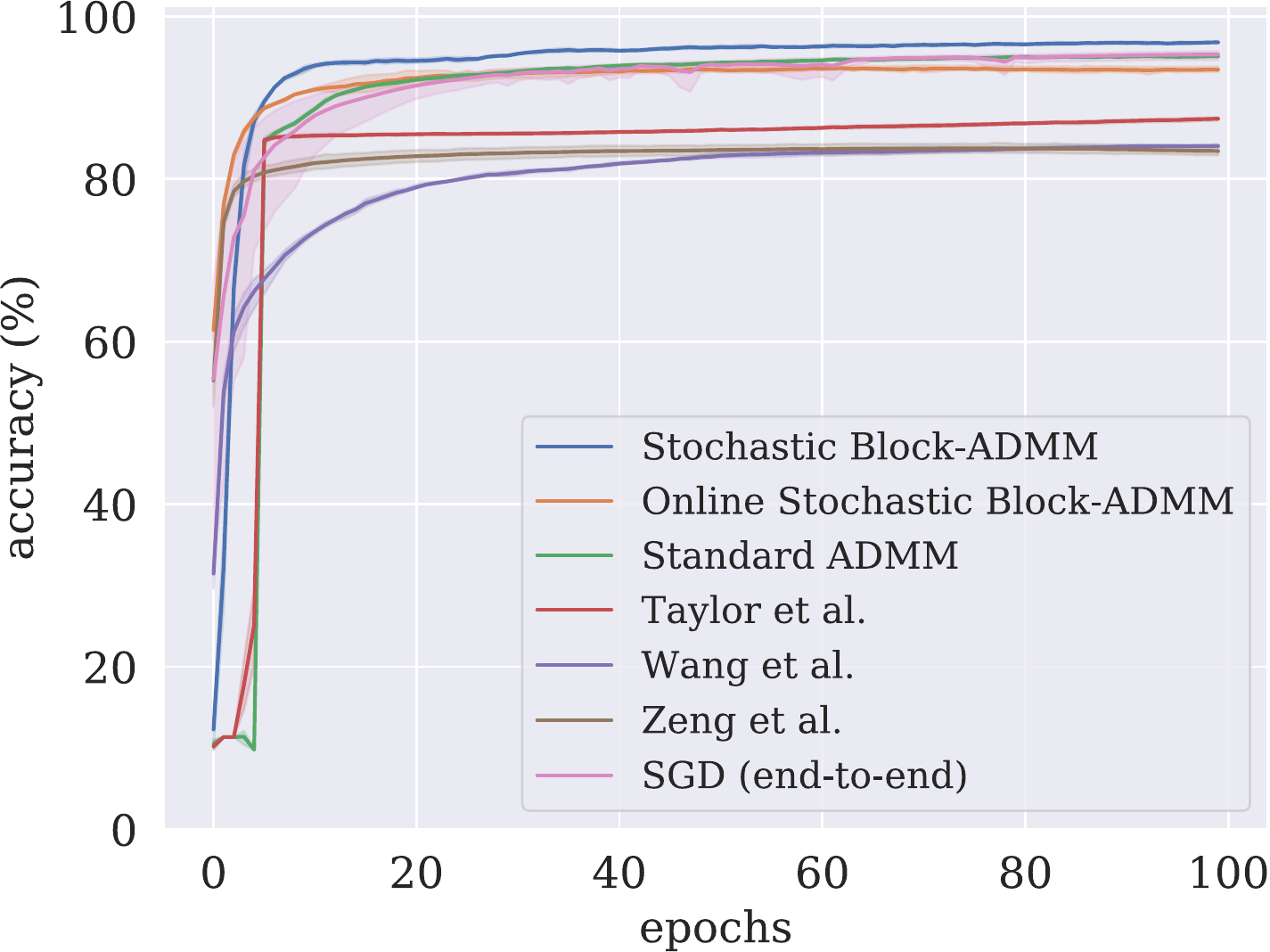}
}
\caption{Test set accuracy on MNIST using network with 3 fully-connected layers: $784-128-128-10$. 
Final test accuracy: ``Stochastic Block-ADMM'': {\bf 97.61\%}, 
``Online Stochastic Block-ADMM'': 93.88\%, 
``Standard ADMM'': 95.02\%, 
\protect \cite{taylor2016training}
: 87.52\%, 
\protect \cite{wang2019admm}: 83.89\% ,
\protect \cite{zeng2018global}: 83.28\% , 
``SGD'': 95.29\% 
(Best viewed in color)}
\label{fig:mnist_acc}
\end{center}
\end{figure}

For the first supervised learning experiment, the MNIST dataset of handwritten digits \cite{mnist}, is used for the evaluation of ADMM/BCD methods for training DNNs. We use the standard train/test split. 
The performance on the testing set of 10,000 samples is reported in Figure \ref{fig:mnist_acc}. The architecture of the \emph{shallow} network used for the experiments incorporates three fully-connected layers with 128-neuron hidden layers $(784-128-128-10)$ and \emph{ReLU} nonlinearity. In order to make a fair comparison with ~\cite{taylor2016training} which can only work with Mean Squared Error (MSE), we utilize MSE as the training objective ($\mathcal{J}$) while the more common Cross-Entropy (CE) is applicable in our block-ADMM formulation and utilized in the experiments in the supplementary materials.

In training standard ADMM and \cite{taylor2016training} as baselines, all the parameters are initialized by sampling from the uniform distribution $x \sim {U}(0, 10^{-4})$.
We set $\beta_l = \gamma_l = 10$ for all of the layers. 
Weight decay is used with $\lambda_l = 5 \times 10^{-5}$. 
For baselines with backpropagation in Fig. \ref{fig:mnist_acc}, a  learning rate of $5 \times 10^{-3}$ is used.

Further, for the training of the batch and online Stochastic Block-ADMM algorithms presented in Algorithm \ref{alg:blockadmm} and \ref{alg:online_admm}, the aforementioned three-layer architecture is split into 3 one-layer blocks. $\beta_t$ is set to 1 for all layers, the weights are initialized using the normal distribution, dual variables $\mU_t$ are initialized using a uniform distribution, and auxiliary variables $\mZ_t$ are initialized in a forward pass. During training, the block parameters ($\Theta_t$) are updated stochastically, and both of sub-problem updates for the $\text{block}_{\Theta_t}$ and $\mZ_t$ are performed using \textit{Adam}. In our experiments in the batch mode, we performed the primal updates for $3$ steps during each iteration. For the online version, we set the batch size to 64 and auxiliary variables are re-initialized at each iteration (see Algorithm \ref{alg:online_admm}). 

Figure \ref{fig:mnist_acc} shows that Stochastic Block-ADMM outperforms the baselines by reaching $97.61 \%$ average test accuracy. Note the accuracy for all methods is lower than normal because of the MSE loss function that is used --- which is not the best choice for classification yet chosen for fair comparison with previous ADMM methods. The online version performs slightly worse with a $93.88 \%$ test accuracy. However, this comes with enormous advantage in terms of memory utilization, e.g. given the configuration for training on MNIST, the online version uses \~ 10$\times$ less memory to store training variables compared to the batch version.

\subsubsection{Vanishing Gradient}\label{exp:vanish}

\begin{figure}[ht]
\begin{center}
\centerline{
\includegraphics[width=\columnwidth]{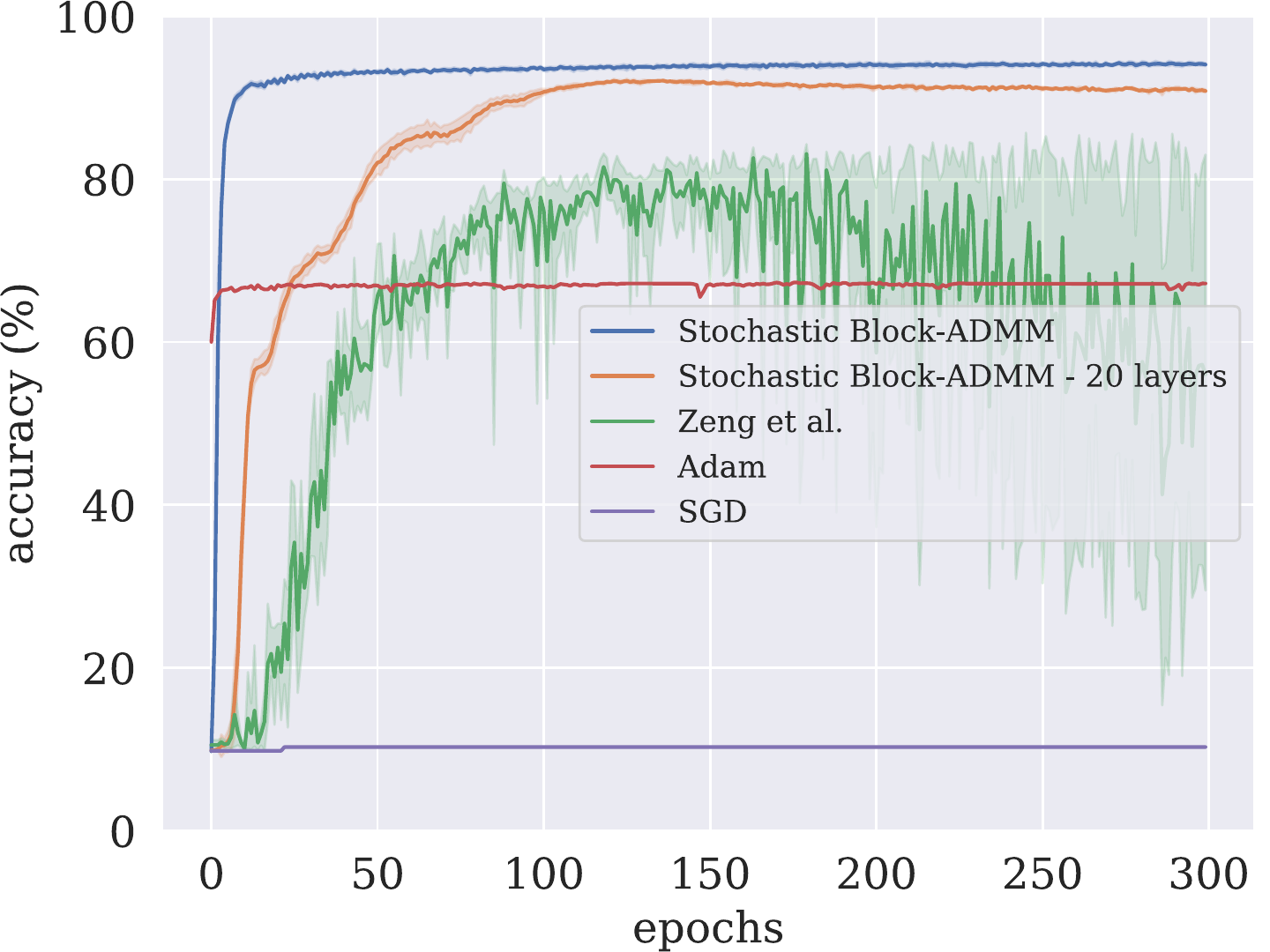}
}
\caption{Test accuracies from deep architectures on MNIST. Block-ADMM demonstrates stable convergence and obtains final test accuracy of $\bf 94.43\%$ (10 layers), and $91.75\%$ (20 layers) respectively, while SGD and Adam (10 layers) fail due to vanishing gradients (Best viewed in color)}
\label{fig:mnist_deep}
\end{center}
\end{figure}

Since no gradient is backpropagated through the entire network in our proposed algorithm, stochastic block-ADMM is robust against vanishing gradients. We run the previous experiments on an unconventional architecture with 10 fully-connected layers --- this is to make the vanishing gradient problem obvious. Note that normally this will not be adopted because of the severe overfitting and gradient vanishing problems, but here we utilized this setting to test our resistance to these problems. Figure \ref{fig:mnist_deep} illustrates the experiment results. Stochastic Block-ADMM reaches final test accuracy of $94.43\%$ while SGD and ADAM only reach to $10.28\%$ and $58\%$, respectively. As it can be seen in Figure \ref{fig:mnist_deep}, we also compared our method with the recent work of \cite{zeng2018global}. We observed the BCD in \cite{zeng2018global} 
to be unstable, sensitive to network architectures, and eventually, not converging after 300 epochs. Although we still exhibited some overfitting, we can see our approach is significantly better in handling of the vanishing gradient problem, and performs reasonably well. We further tested our performance with 20 fully-connected layers. Results show that although there is slightly more overfitting, our algorithm can still find a reasonable solution (Fig.~\ref{fig:mnist_deep}), showing its potential in helping with training scenarios with vanishing gradients.

\subsubsection{Wall Clock Time Comparison}\label{time_cmp}

\begin{figure}[ht]
\begin{center}
\centerline{
\includegraphics[width=\columnwidth]{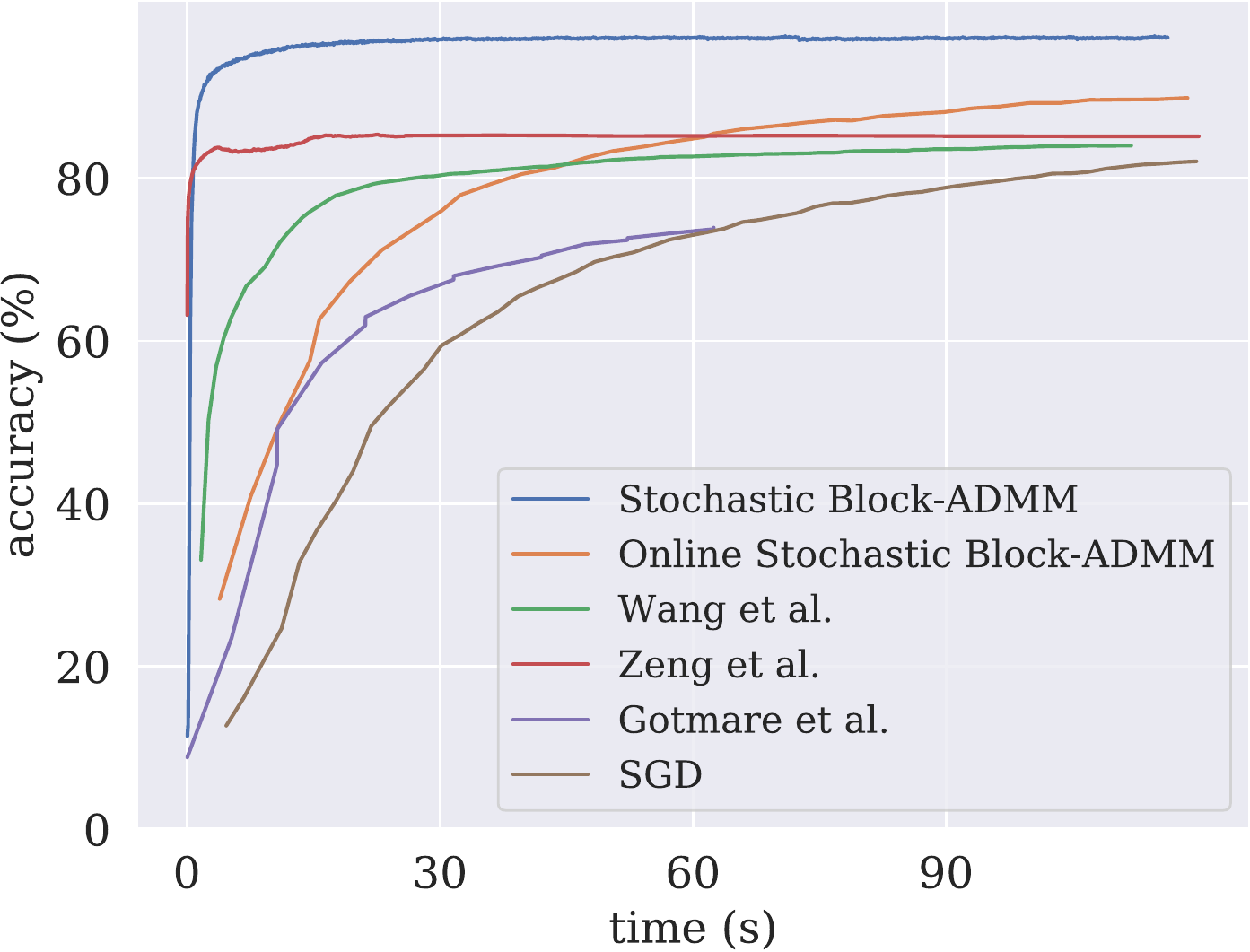}
}
\caption{Test set accuracy v.s. training wall clock time comparison of different alternating optimization methods for training DNNs on the MNIST dataset. Our methods (blue and orange) show superior performance vs. \protect\cite{zeng2018global} and \protect\cite{wang2019global} while converge faster than all other methods}
\label{fig:time}
\end{center}
\end{figure}

In this section, we analyze the batch and online versions of stochastic block-ADMM in training wall clock time and compare them against other baselines as illustrated in Figure \ref{fig:time}.  
Note Gotmare \etal and SGD are trained with a mini-batch size of 64 and \cite{zeng2018global,wang2019admm} are trained in a batch setting. Only the time taken for the \emph{training} was plotted in Fig.~\ref{fig:time} and stages such as initialization, data loading, etc were excluded. The online version shows faster convergence than \cite{gotmare2018decoupling} and simple SGD. Although \cite{zeng2018global} and \cite{wang2019global} have been convergence rates due to being batch methods, our approach achieves higher performance later on.


\subsection{Supervised Disentangling on LFWA}\label{exp:hetero}

In this section, we showcase the flexibility of stochstic block-ADMM in trainig deep networks with non-differentiable layers where conventional backpropagation cannot be used. For that purpose, we evaluate our proposed method in a supervised disentanglement problem where we used DeepFacto \ref{sec:deepfacto} to learn a nonnegative factorized representation of the DNN activations while training end-to-end on the LFWA dataset \cite{LFWTech}. Next, similar to \cite{liu2018exploring}, linear SVMs are used over the factorized space to predict face attributes. This setup examines the capability of the network to extract a disentangled representation that linearly corresponds to human-marked attributes that the network does not have prior knowledge of.

We used the Inception-Resnet architecture from \cite{schroff2015facenet}, pre-trained on the VGGFace-2 \cite{Cao18} dataset as the back-bone. To incorporate an NMF, we follow the same approach as in Fig.~\ref{fig:deepfacto} where the pretrained DNN is the first block, and we add a simple fully-connected layer over the score matrix $\mS_t$ to train a face-verification network with a triplet loss~\cite{hoffer2015deep}.
We conjecture the score matrix $\mS_t$ will be guided to learn an disentangled factorization due to the nonnegativity constraint \cite{collins2018deep}. 
To have a warm start for an end-to-end training of DeepFacto, we first pre-train the NMF module having the Inception-Resnet block freezed. Then, we fine-tune the block parameters as well as the NMF module in an alternating fashion, similar to Algorithm \ref{alg:blockadmm}. Note, the rank of the NMF in DeepFacto is a hyperparameter and we selected three different values ($r=4, 32, 256$) in the experiments. The final $r=256$ is also the latent space dimensionality in \cite{liu2018exploring}.
Table. \ref{table:lfw} illustrates average prediction accuracy over LFWA attributes
from DeepFacto and other supervised and weakly supervised baselines. This validates that DeepFacto has learned a meaningful representation of the attributes by disentangling the activations. To see visualization for individual dimensions learned by DeepFacto see supplementary materials \ref{sec:weakly_sup}.


\begin{table}[t]
\caption{Average prediction accuracy on 40 attributes from LFWA dataset. Weakly-supervised methods train the network without access to attribute labels. Final classification then comes from a linear SVM on their latent representations.}
\label{table:lfw}
\begin{center}
\begin{small}
\begin{sc}
\begin{tabular}{lcccr}
\toprule
LFWA & Accuracy \\
\midrule
\cite{zhang2014panda} {\tiny (supervised)}                      &  81.00\%\\
\cite{liu2015deep} {\tiny (supervised)}                         &  84.00\%\\
\cite{liu2018exploring}  {\tiny (weakly-supervised)}             &  83.16\%\\
Deepfacto - rank 4 {\tiny (weakly-supervised)}                   & 74.80\%\\
Deepfacto - rank 32 {\tiny (weakly-supervised)}                  & 81.39\%\\
Deepfacto - rank 256 {\tiny (weakly-supervised)}                 & \textbf{87.03}\%\\
\bottomrule
\end{tabular}
\end{sc}
\end{small}
\end{center}
\end{table}


\section{Conclusion and Discussion}
In this paper, we proposed stochastic block-ADMM as an approach to train deep networks. Through updates with stochastic gradients, we improve over the capabilities to scale to larger networks using ADMM, as well as the performance. We alps presented an online version of stochastic block-ADMM for setting where computational power is limited, or when accessing to all data at once is not practical. 
We have shown improvements over SGD/Adam in training deep networks without residual connections. As an illustration to how ADMM can be applied in supervised feature disentanglement, we propose DeepFacto which jointly trains an NMF layer within a deep network and show encouraging results on a supervised disentanglement benchmark, both quantitatively and qualitatively. We believe the results presented in this work set up future work that further explores aspects of utilizing ADMM in deep network training, including parallelization and stability.

\bibliographystyle{style}
{\small
\bibliography{ref}
}

\clearpage
\appendix 
\section*{Supplementary Materials}
\section{Background: Standard ADMM Training of DNNs} \label{sec:admm_nn}

Alternating Direction Method of Multipliers (ADMM) \cite{gabay1975dual,boyd2011distributed} is a class of optimization methods belonging to  \textit{operator splitting techniques} which borrows benefits from both dual decomposition and augmented Lagrangian methods for constrained optimization. 

To formulate training an $L$-layer DNN in a general supervised setting, we would have the following non-convex constrained optimization problem \cite{zeng2018global}:
\begin{align} \label{eq:obj}
	\minimize_{ \mathcal{W}, \mathcal{A}, \mathcal{Z}} \quad &\mathcal{J}\left(\mY, \mZ_{L} \right) + \sum_{\ell = 1}^{L} \lambda_{\ell}  {\bf r}_{\ell} (\mW_{\ell}) \\
	 {\rm subject~to} \quad & \mA_{\ell} - {\bm \phi}_{\ell } \left( \mZ_{\ell} \right) = {\bf 0}, \quad \ell = 1,\dots, L-1   \nonumber \\
	 {\rm subject~to} \quad & \mZ_{\ell} - \mW_{\ell} \mA_{\ell-1} = {\bf 0}, \quad \ell = 1, \dots , L \nonumber 
\end{align}
where $\mathcal{J}$ is the main objective (\textit{e.g.}, cross-entropy, mean-squared-error loss functions) that needs to be minimized. The subscript $\ell$ denotes the $\ell$-th layer in the network. The optimization variables are $\mathcal{W} = \{ \mW_\ell\}_{\ell=1}^{L}$, $\mathcal{A} = \{ \mA_{\ell}\}_{\ell=1}^{L-1}$, and $\mathcal{Z} = \{ \mZ_{\ell}\}_{\ell=1}^{L}$ where $\mW_\ell$, $\mZ_{\ell}$, $\mA_\ell$, and ${\bm \phi}_\ell (.)$ are the weight matrix, output matrix, activation matrix, and the activation function (\textit{e.g.}, ReLU) at the $\ell$-th layer, respectively. Note that $\mA_{0} = \mX$ where $\mX = \{ \vx_1,\dots, \vx_N \} \in  \R^{M \times N}$ is the input data matrix containing $N$ samples with input dimensionality $M$; $\mY = \{\vy_1,\dots, \vy_N \} \in \R^{C \times N}$ is the target matrix pair comprised of $N$ one-hot vector label of dimension $C$, representing number of prediction classes. Also, ${\bf r(.)}$ is the regularization term with (\textit{e.g.}, Frobenius norm $\|.\|_F^2$) corresponding penalty weight $\lambda_{\ell}$. Note that the regularization term can be simply ignored by setting $\lambda_\ell$ to zero. In this formulation, the intercept in each layer is ignored for simplicity as it can be simply be added by slightly modifying the $\mW_\ell$ and the input to each layer. The formulation in Eq. (\ref{eq:obj}) breaks the the conventional multi-layer backpropagation optimization of DNNs into simpler sub-problems that can be solved efficiently (e.g. reducing to least-squares problem). This also facilitates training in a distributed manner --- as the layers of the DNN are decoupled and the variables can be updated in parallel across layers ($\mW_\ell$) and data points (\ $\mW_\ell, \mZ_\ell, \mA_\ell$).

To enforce the constraints in problem (\ref{eq:obj}) and solve the optimization using ADMM, we would have the following augmented Lagrangian problem:

\begin{eqnarray} \label{eq:augmented}
	\minimize_{ \mathcal{W}, \mathcal{A}, \mathcal{Z}} \quad &\mathcal{J}\left(\mY, \mZ_{L} \right) + \sum_{\ell = 1}^{L} \lambda_{\ell}  {\bf r}_{\ell} (\mW_{\ell}) \\
	& + \sum_{\ell=1}^{L} \frac{\beta_{\ell}}{2} \| \mZ_{\ell} - \mW_{\ell} \mA_{\ell-1} + \mU_{\ell}\|_{F}^{2} \nonumber\\
	& + \sum_{\ell=1}^{L-1} \frac{\gamma_{\ell}}{2} \| \mA_{\ell} - {\bm \phi}_{\ell}(\mZ_{\ell}) + \mV_{\ell}\|_{F}^{2}\nonumber
\end{eqnarray}
where $\beta_{\ell}, \gamma_\ell >0$ are the step sizes, $\mU_{\ell}$ and $\mV_{\ell}$ are the \textit{(scaled) dual variables} \cite{boyd2011distributed} for the equality constraint at the layer $\ell$. 
Algorithm \ref{alg:admm} shows a standard ADMM scheme for optimizing Eq. (\ref{eq:augmented}). Note, the parameters are updated in a closed-form as analytical solution can be simply derived. For simplicity of the equations, we denote $\gP_\ell (.) = \frac{\beta_{\ell}}{2} \| \mZ_{\ell} - \mW_{\ell} \mA_{\ell-1} + \mU_{\ell}\|_{F}^{2} $ and $\gQ_\ell (.) = \frac{\gamma_{\ell}}{2} \| \mA_{\ell} - {\bm \phi}_{\ell}(\mZ_{\ell}) + \mV_{\ell}\|_{F}^{2}$. This algorithm is similar to \cite{taylor2016training,wang2019admm} with the difference that all the equality constraints in problem (\ref{eq:obj}) are enforced using multipliers, while previous work only enforced the constraints on the last layer $L$ while other constraints were only loosely enforced using quadratic penalty. 

\begin{algorithm}[htb]
  \caption{Standard ADMM for DNN Training}
  \label{alg:admm}
\begin{algorithmic}
  {\STATE \scalebox{1}{\bfseries Input:} data $\mX$, labels $\mY$}
  \STATE  \scalebox{1}{{\bfseries Params:} $\beta_\ell >0, \gamma_\ell >0,\lambda_\ell > 0$ }
  \STATE  \scalebox{0.8}{{\bfseries Initialize:} $\{\mW_\ell^0\}_{\ell=1}^{L}, \{ \mU_\ell^0\}_{\ell=1}^{L}, \{ \mV_\ell^0\}_{\ell=1}^{L-1}, \{\mZ^0_\ell\}_{\ell=1}^{L}, \{\mA^0_\ell\}_{\ell=1}^{L-1}\; k \leftarrow 0$ }
  \REPEAT
  \FOR{$\ell=1$ {\bfseries to} $L$}
  \STATE \scalebox{1}{$\mW_\ell^{k+1} \leftarrow \argmin\; \{ \gP_\ell (.) +  \lambda_{\ell}  {\bf r}_{\ell} (\mW_{\ell}^{k})\}$}
  \ENDFOR
  \FOR{$\ell=1$ {\bfseries to} $L-1$}
  \STATE \scalebox{1}{ $\mZ_\ell^{k+1} \leftarrow \argmin\; \{ \gP_\ell (.) +  \gQ_\ell (.) \}$ }
  \STATE \scalebox{1}{$\mA_\ell^{k+1} \leftarrow \argmin\; \{ \gP_{\ell+1} (.) +  \gQ_\ell (.) \} $}
  \ENDFOR
    \STATE \scalebox{1}{ $\mZ_{L}^{k+1} \leftarrow \argmin\; \{ \mathcal{J}\left(\mY, \mZ_{L}^{k} \right) + \gP_L (.) \}$ }
  \FOR{$\ell=1$ {\bfseries to} $L-1$}
  \STATE \scalebox{1}{$\mU_\ell^{k+1} \leftarrow \mU_\ell^{k} + \mZ_{\ell}^{k+1} - \mW_{\ell}^{k+1} \mA_{\ell-1}^{k+1}$}
  \STATE \scalebox{1}{$\mV_\ell^{k+1} \leftarrow \mV_\ell^{k} + \mA_{\ell}^{k+1} - {\bm \phi}_{\ell}(\mZ_{\ell}^{k+1})$}
  \ENDFOR
  \STATE \scalebox{1}{$\mU_L^{k+1} \leftarrow \mU_L^{k} + \mZ_{L}^{k+1} - \mW_{L}^{k+1} \mA_{L-1}^{k+1}$}
  \UNTIL{some stopping criterion is reached.}
\end{algorithmic}
\end{algorithm}

While the standard ADMM Algorithm \ref{alg:admm} has potentials in training (simple) DNNs \cite{taylor2016training}, there exists hurdles that confines extending ADMM to more complex problems --- the global convergence proof of the ADMM \cite{deng2016global} assumes that $\mathcal{J}$ is deterministic and the global solution is calculated at each iteration of the cyclic parameter updates.
This makes standard ADMM computationally expensive thus impractical for training of many large-scale optimization problems. Specifically, for  deep learning, this would impose a severe restriction on training set size when limited computational resources are available. In addition, since the variable updates in standard ADMM are analytically driven, the extent of its applications is limit to trivial tasks \cite{taylor2016training}, making it incompetent to perform on par with the recent complex architectures introduced in deep learning (e.g. \cite{he2016deep}).

\section{Proof for Proposition 1}\label{sec:proof}

We follow the steps in the proof for similar problems in \cite{fu2018anchor} and \cite{shi2017penalty} with deterministic primal updates. Proper modifications are made to cover the stochastic primal update in our proof.

Note that we have
              \[     \nabla{\cal L}_{\rho_k}(\X^k)= \nabla f(\X^k) + \nabla h(\X^k)^T\bm \mu^k,          \]
              where 
              \[      \bm \mu^k = (1/\rho_k)h(\bm X^k)+\bm \lambda^k.   
              \]
              Our first step is to show that $\{\bm \mu^k\}$ is a convergent sequence. To see this, we define 
              \[ \bm \bar{\bm \mu}^k = \frac{\bm \mu^k}{\|{\bm \mu}^k\|}. \]
              Since $\bm \bar{\bm \mu}^k$ is bounded, it converges to a limit point $\bm \bar{\bm \mu}$. Also let $\x^\star$ be a limit point of $\x^k$.
              Because we have assumed that 
              $$\varepsilon_k\rightarrow 0,\quad \sigma_k^2\rightarrow 0,$$ 
              it means that the mean and variance of the stochastic gradient of our primal update goes to zero.
              Since our stochastic gradient is unbiased, we have
              \[       {\cal G}(\X^k) \rightarrow \nabla {\cal L}_{\rho_k}(\X^\star). \]  
              This also means that  we must have ${\cal G}(\x^k)\rightarrow \bm 0$ and $$\nabla L_{\rho_k}(\bm x^k)\rightarrow \bm 0.$$
     Hence, the following holds when $k\rightarrow \infty$:
              \begin{equation}\label{eq:approxkkt}
                 \nabla L_{\rho_k}(\bm X^\star)=\nabla f(\X^\star)+\nabla h(\X^\star)^T\bm {\bm \mu}^\infty = 0,
              \end{equation}

              Suppose $\bm \mu^k$ is unbounded. By dividing \eqref{eq:approxkkt} by the above $\|\bm \mu^k\|$ and considering $k\rightarrow \infty$, we must have 
              \begin{equation}\label{eq:key}
                \nabla h(\X^\star)^T\bm \bar{\bm \mu}= 0,\quad \forall \X.    
              \end{equation}               
              The term $\nabla f(\bm X^\star)/\|\bm \mu\|$ is zero since we assumed $\bar{\bm \mu}$ is unbounded.
              Since $h(\bm X)=\bm 0$ satisfies the Robinson's condition, then, for any $\bm w$, there exists $\beta>0$ and $\bm x$ such that
              \[      \bm w = \beta \nabla h(\X^\star)(\X-\X^\star).        \]
              This together with \eqref{eq:key} says that $\bar{\bm \mu}=\bm 0$. This contradicts to the fact $\|\bar{\bm \mu}\|=1$. Hence, $\{ \bm \mu^k \}$ must be a bounded sequence and thus admits a limit point. Denote $\bm \mu^\star$ as this limit point, and take limit of both sides of \eqref{eq:approxkkt}. We have:
              \begin{equation}
              \nabla f(\X^\star)+\nabla h(\X^\star)^T\bm \mu^\star= \bm 0,\quad \forall \X.
              \end{equation}
               
              In addition, since $$\rho_k(\bm \mu^k-\bm \lambda^k) = h(\mathbf{\X^k})$$ with $\rho_k \rightarrow 0$ or $\bm \mu_k-\bm \lambda_k \rightarrow 0$ (per our updating rule and $\eta_k\rightarrow 0$), the constraints will be enforced in the limit.      $\mbox{     } \square$   \\




\begin{figure}[ht]
\begin{center}
\centerline{
\includegraphics[width=\columnwidth]{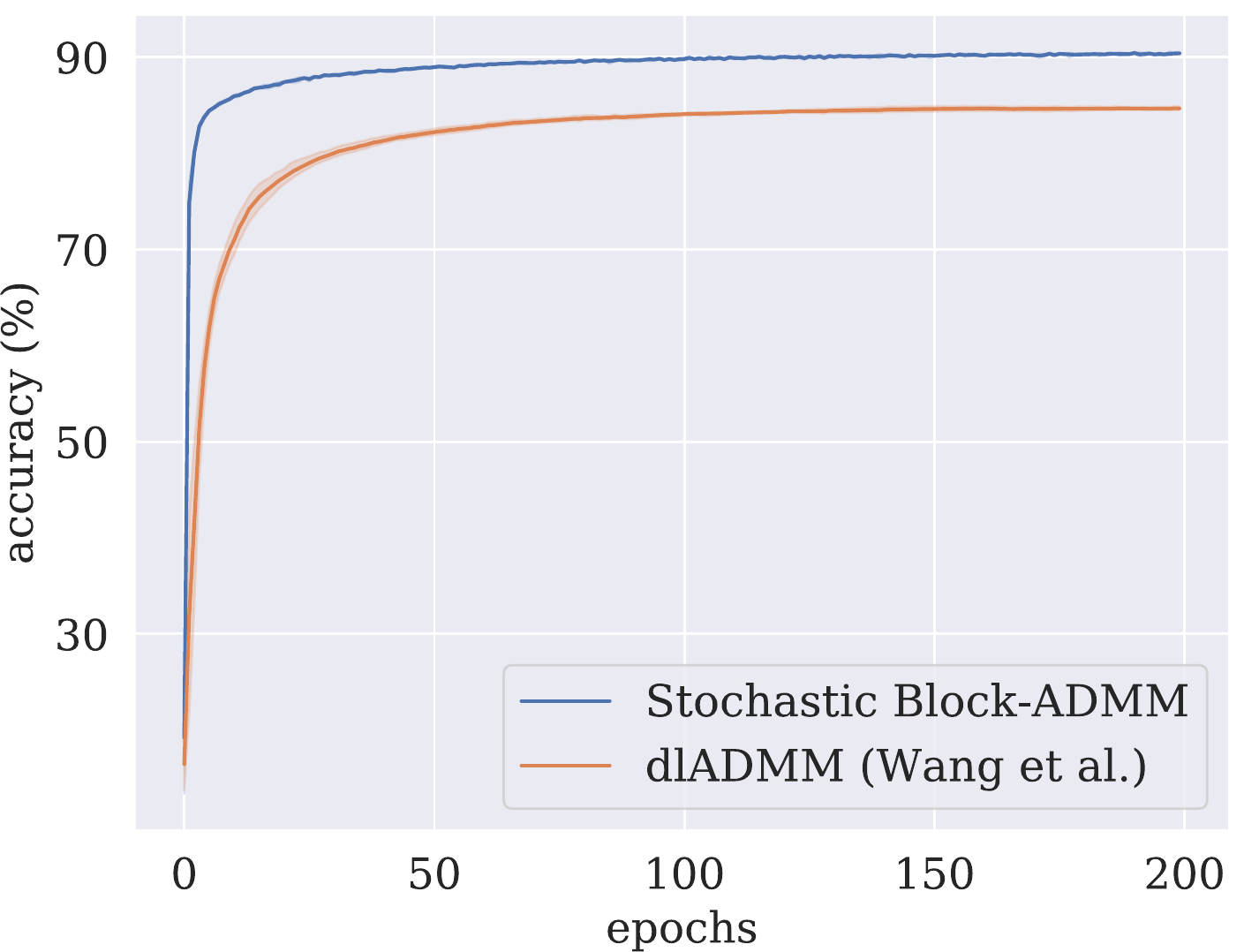}
}
\caption{Test accuracy comparison of Stochastic Block ADMM and dlADMM on Fashion-MNIST dataset using a network with 3 fully-connected layers: $784-1000-1000-10$. Final test accuracy: "Stochastic Block ADMM": $\bf 90.39\%$, "Wang \textit{et al.}":$84.67 \%$ (averaged over 5 runs).}
\label{fig:fmnist_acc}
\end{center}
\end{figure}

\section{Supervised training of DNNs}\label{sec:sup_train}

\textbf{Fashion-MNIST.}
To compare our method with dlADMM \cite{wang2019admm}, we evaluated the performance of our method on the Fashion-MNIST dataset \cite{xiao2017/online} with 60,000 training samples and 10,000 testing samples. We followed the settings in \cite{wang2019admm} by having 2 hidden layers with 1000 neurons each, and Cross-Entropy loss at the final layer. Also, the batch size is set to 128, $\beta_t = 1$, and the updates for $\mZ_t$ and $\Theta_t$ (eq. 6a) are performed 3 times at each epoch. Figure \ref{fig:fmnist_acc} shows the test set accuracy results over 200 epochs of training. It can be noticed that Stochastic Block ADMM is converging at lower epochs and reaching a higher test accuracy while performing efficient mini-batch updates. Further, in section C., it will be demonstrated that Stochastic Block ADMM converges drastically faster than dlADMM in terms of wall clock time.

\textbf{CIFAR-10.}
The previous works on training deep netowrks using ADMM have been limited to trivial networks and datasets (e.g. MNIST) \cite{taylor2016training,wang2019admm}. However, our proposed method does not have many of the existing restrictions and assumptions in the network architecture, as in previous works do, and can easily be extended to train non-trivial applications. It is critical to validate stochastic block-ADMM in settings where deep and modern architectures such as deep residual networks, convolutional layers, cross-entropy loss function, etc., are used. To that end, we validate the ability of our method is a supervised setting (image classification) on the CIFAR-10 dataset \cite{cifar} using ResNet-18 \cite{he2016deep}. To best of our knowledge, this is the first attempt of using ADMM for training complex networks such as ResNets.

For this purpose, we used 50,000 samples for training and the remaining 10,000 for evaluation. 
To have a fair comparison, we followed the configuration suggested in \cite{gotmare2018decoupling} by converting Resnet-18 network into two blocks $(T=2)$, with the splitting point located at the end of {\sc conv3\_x} layer. We used the Adam optimizer to update both the blocks and the decoupling variables with the learning rates of $\eta_t = 5e^{-3}$ and $\zeta_t = 0.5$. We noted since the auxiliary variables $\mZ_t$ are not "shared parameters" across data samples, they usually require a higher learning rate in Algorithm \ref{alg:blockadmm}. Also, we found the ADMM step size $\beta_t = 1$ to be sufficient for enforcing the block's coupling.

\begin{figure}[htb]
\begin{center}
\centerline{
\includegraphics[width=\columnwidth]{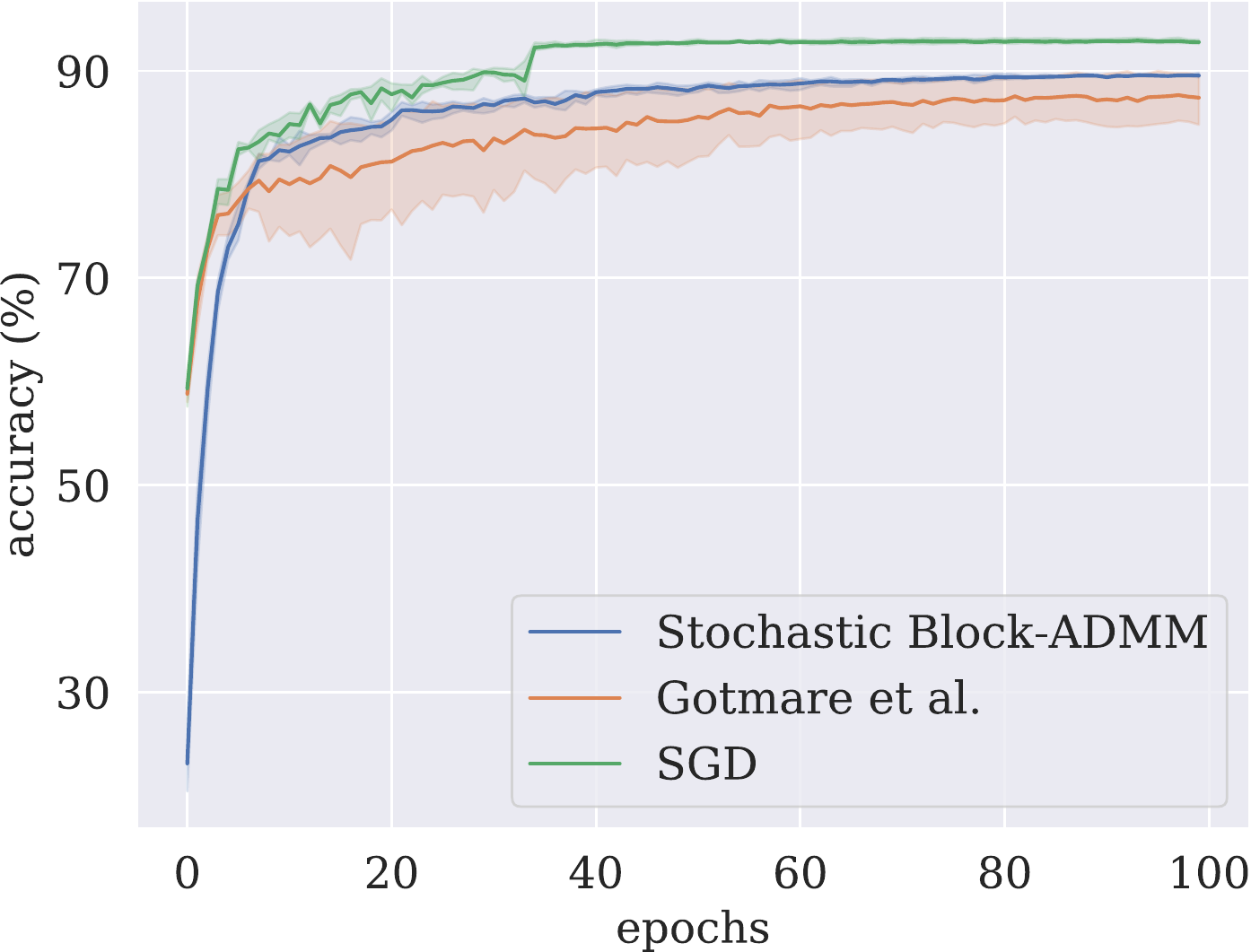}
}
\caption{Test set accuracy on CIFAR-10 dataset. Final accuracy "Block ADMM": $89.66\%$, "Gotmare \etal":$87.12 \%$, "SGD": $\bf 92.70\%$. (Best viewed in color.)}
\label{fig:cifar}
\end{center}
\end{figure}

Figure. \ref{fig:cifar} shows the results from our method compared with two baselines: \cite{gotmare2018decoupling}, and conventional end-to-end neural network training using back-propagation and SGD. Our algorithm consistently outperformed ~\cite{gotmare2018decoupling} however cannot match the conventional SGD results. There are several factors that we hypothesize that might have contributed to the performance difference: 1) in a ResNet the residual structure already partially solved the vanishing gradient problem, hence SGD/Adam performs significantly better than a fully-connected version; 
2) we noticed decreasing the learning rate for $\Theta_t$ updates does not impact the performance as it does for an end-to-end back-propagation using SGD. Still, we obtained the best performance of ADMM-type methods on both MNIST and CIFAR datasets, showing the promise of our approach.


\begin{table*}[htb]
\caption{Prediction accuracy (\%) of individual attributes in LFWA dataset. DeepFacto with other weakly-supervised and supervised baselines.}
\label{table:attr_lfw}
\vskip 0.15in
\begin{center}
\begin{small}
\begin{sc}
\begin{tabular}{lcccccc}
\toprule
{Attributes} & \multicolumn{3}{c}{\small DeepFacto} & \small \cite{liu2015deep} & \small \cite{liu2018exploring} & \small \cite{zhang2014panda}\\
 {} & \multicolumn{3}{c}{\tiny (Weakly-Supervised)} & {\tiny (Weakly-Supervised)} & {\tiny (Supervised)} & {\tiny (Supervised)} \\
 {} & $r= $256 & 32 & 4 \\
\midrule
‘5 o Clock Shadow’ & 83.3 & 80.0 & 68.7 & 78.8 & \bf84 & \bf84\\
‘Arched Eyebrows’ & \bf86.6 & 83.9 & 79.2 & 78.1 & 82 & 79\\
‘Attractive’ & \bf84.3 & 79.8 & 73.3 & 79.2 & 83 & 81\\
‘Bags Under Eyes’ & \bf83.9 & 72.5 & 64.5 & 83.1 & 83 & 80 \\
‘Bald’ & \bf94.3 & 93.3 & 89.3 & 84.8 & 88 & 84\\
‘Bangs’ & \bf93.2 & 88.4 & 84.4 & 86.5 & 88 & 84\\
‘Big Lips’ & \bf83.2 & 77.0 & 71.9 & 75.2 & 75 & 73\\
‘Big Nose' & 80.1 & 68.7 & 61.4 & \bf81.3 & 81 & 79\\
‘Black Hair’ & \bf92.7 & 91.4 & 87.4 & 87.4 & 90 & 87\\
‘Blond Hair’ & \bf97.9 & 97.3 & 93.2 & 94.2 & 97 & 94\\
‘Blurry’ & \bf90.4 & 90.5 & 86.5 & 78.4 & 74 & 74\\
‘Brown Hair’ & \bf78.4 & 74.4 & 70.2 & 72.9 & 77 & 74\\
‘Bushy Eyebrows’ & \bf84.0 & 78.6 & 63.4 & 83.0 & 82 & 79\\
‘Chubby’ & \bf80.5 & 75.2 & 71.1 & 74.6 & 73 & 69\\
‘Double Chin’ & \bf86.0 & 77.9 & 72.3 & 80.2 & 78 & 75\\
‘Eyeglasses’ & 94.3 & 89.6 & 84.8 & 89.5 & \bf95 & 89\\
‘Goatee’ & \bf89.1 & 85.4 & 80.0 & 78.6 & 78 & 75\\
‘Gray Hair’ & \bf91.9 & 90 & 85.6 & 86.9 & 84 & 81\\
‘Heavy Makeup’ & \bf96.3 & 91.5 & 87.4 & 94.5 & 95 & 93\\
‘High Cheekbones’ & \bf90.4 & 79.0 & 72.1 & 88.8 & 88 & 86\\
‘Male’ & 81.3 & 76.6 & 70.5 & \bf94.3 & 94 & 92\\
‘Mouth Slightly Open’ & \bf85.4 & 78.0 & 73.3 & 81.7 & 82 & 78 \\
‘Mustache’ & \bf96.6 & 93.2 & 91.3 & 83.3 & 92 & 87\\
‘Narrow Eyes’ & \bf78.3 & 69.3 & 58.4 & 77.5 & 81 & 73\\
‘No Beard’ & \bf79.5 & 73.0 & 65.5 & 77.7 & 79 & 75\\
‘Oval Face’ & \bf80.6 & 73.2 & 66.1 & 78.7 & 74 & 72\\
‘Pale Skin’ & 75.1 & 66.7 & 60.6 & \bf89.8 & 84 & 84\\
‘Pointy Nose'& \bf81.6 & 73.7 & 62.2 & 79.8 & 80 & 76\\
‘Receding Hairline’ & 84.0 & 80.9 & 73.8 & \bf88.0 & 85 & 84 \\
‘Rosy Cheeks’ & \bf87.3 & 87.4 & 83.4 & 79.9 & 78 & 73\\
‘Sideburns’ & \bf85.4 & 81.5 & 75.8 & 80.5 & 77 & 76\\
‘Smiling’ & \bf92.6 & 78.7 & 69.8 & 92.2 & 91 & 89\\
‘Straight Hair’ & \bf82.8 & 77.0 & 72.1 &  73.6 & 76 & 73\\
‘Wavy Hair’ & 80.4 & 77.0 & 68.3 & \bf81.7 & 76 & 75\\
‘Wearing Earrings’ & \bf95.4 & 91.6 & 87.1 & 89.7 & 94 & 92\\
‘Wearing Hat’ & \bf93.0 & 90.2 & 87.0 & 80.5 & 88 & 82\\
‘Wearing Lipstick’ & \bf95.8 & 92.8 & 89.0 & 91.4 & 95 & 93\\
‘Wearing Necklace’ & \bf93.0 & 89.8 & 85.1 & 84.0 & 88 & 86\\
‘Wearing Necktie’ & \bf79.8 & 75.2 & 70.6 & 78.7 & 79 & 79\\
‘Young’ & \bf91.0 & 88.4 & 84.4 & 79.2 & 86 & 82\\
\midrule
Average & \bf87.0 & 81.4 & 74.8 &  83.1 & 84 & 81\\
\bottomrule
\end{tabular}
\end{sc}
\end{small}
\end{center}
\vskip -0.25in
\end{table*}

\section{Weakly Supervised Attribute Prediction}\label{sec:weakly_sup}

\subsection*{Factorizing the activations}\label{sec:factor_layer} 

With the assumption that the observations are formed by a linear combination of few basis vectors, one can approximate a given matrix $\mX \in \R^{m \times n}$ into a \textit{basis} matrix $\mM \in \R^{m \times r}$ and an \textit{score} matrix $\mS \in \R^{r \times n}$ such that $\mX \approx \mM \mS$ where $r$ is the (reduced) \textit{rank} of the factorized matrices -- commonly $r \ll \min(m, n)$.
Methods such as NMF would restrict the entries of $\mM$ and $\mS$ to be non-negative $(\forall i,j \;  \mM_{ij} \ge 0,\; \mS_{ij} \ge 0)$ which forces the decomposition to be only \textit{additive}. This has been shown to result in a parts-based representation that is intuitively more close to human perception. It is also worth mentioning that obviously, the matrix $\mX$ needs to be positive $({\forall i,j} \;  \mX_{ij} \ge 0)$. For non-negative factorization on the activations of the DNNS, due to the common use of activation functions such as \textit{ReLU}, this would not impose any constraints in most of the problems.

Activations of the CNN networks are generally tensors of the shape $\tZ_{\ell} \in \R^{(N, C, H, W)}$ which namely represent the batch size of the input, the number of the channels, the height of each channel, and the corresponding width. To adapt such tensors for the NMF problem, we reshape the tensor into the matrix $\mZ_{\ell} \in \R^{ C \times (N * H * W)}$ by stacking it over its channels while flattening the other dimensions. This way, the channels would be embedded into a pre-defined small dimension $r$ while keeping each sample and pixels information. For the weakly-supervised problem of attribute classification using DeepFacto, we attached the DeepFacto module to the last convolutional layer of the Inception-Resnet-V1 architecture followed by a \emph{ReLU}. This layer has 1792 channels and, for a given input of the size $160 \times 160$ pixels (the original input size from the LFWA dataset), the height and the width are both equal to 3. 

\begin{figure}[htb]
\vskip -0.05in
\begin{center}
\centerline{
\includegraphics[width=\columnwidth]{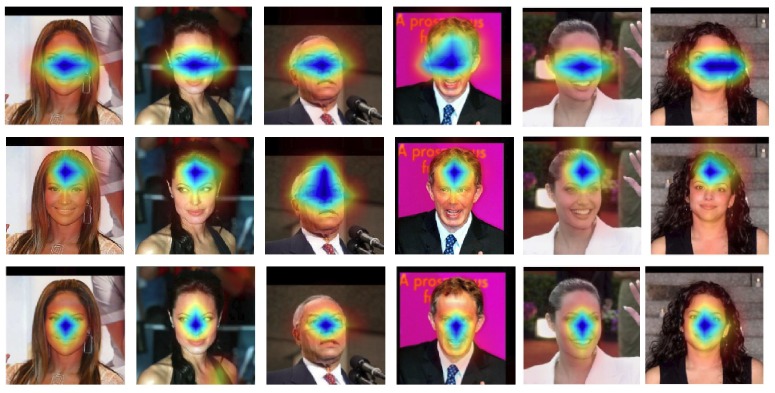}
}
\caption{Heat map visualizations from three different dimensions of the score matrix $\mS$ (rows) trained by DeepFacto-32 over different samples (columns) in LFWA dataset. These dimensions can capture interpretable representations over different faces identities: \emph{eyes} (top), \emph{forehead} (middle), and \emph{nose} (bottom).}
\label{fig:heatmap}
\end{center}
\vskip -0.15in
\end{figure}


\subsection*{Heat maps}\label{sec:heatmap}
To qualitatively investigate the interpretability of the factorized representations learned from DeepFacto, similar to \cite{collins2018deep}, one can visualize the score matrix $\mS$. Each dimension of the score matrix $\mS$ can be reshaped back to the original activation size and be up-sampled to the size of the input using bi-linear interpolation. In Figure \ref{fig:heatmap}, the score matrix learned form the DeepFacto with $r=32$ (average attribute prediction of 81.4\%) is used where three different heat maps (out of 32) are depicted over different samples from LFWA dataset. We have found $r=4$ to be very low to represent interpretable heat maps for the attributes and $r=256$ to contain redundant heat maps. It can be seen, that the heat maps can show local and persistent attention over different face identities: \emph{eyes}, \emph{forehead}, \emph{nose}, etc.


\end{document}